\documentclass[hidelinks]{article}
\usepackage[preprint]{nips_2018}
\usepackage{caption} 
\usepackage{wrapfig}
\usepackage{float}
\usepackage[utf8]{inputenc} 
\usepackage[T1]{fontenc}    
\usepackage{hyperref}       
\usepackage{url}            
\usepackage{booktabs}       
\usepackage{amsfonts}       
\usepackage{nicefrac}       
\usepackage{microtype}      
\usepackage{graphicx}
\usepackage{amsmath}
\usepackage{natbib}
\usepackage{dsfont}
\usepackage{xcolor}
\usepackage{subcaption}
\usepackage{verbatim}
\usepackage{enumitem}
\usepackage{titlesec}
\usepackage{tabularx}
\usepackage[para]{footmisc}

\DeclareMathSizes{10}{9}{7}{5}
\titlespacing\section{0pt}{12pt plus 4pt minus 2pt}{0pt plus 2pt minus 2pt}
\setcitestyle{numbers}
\setcitestyle{square}

\title{One-Shot Federated Learning}
\author{
Neel Guha\\
Carnegie Mellon University \\
\texttt{neelguha@gmail.com} \\
\And
Ameet Talwalkar\\
Carnegie Mellon University \\
\texttt{talwalkar@cmu.edu} \\
\And
Virginia Smith\\
Carnegie Mellon University \\
\texttt{smithv@cmu.edu} \\
}

\begin{document}

\setlength{\abovedisplayskip}{0pt}
\setlength{\belowdisplayskip}{3pt}
\maketitle
\begin{abstract}
We present \textit{one-shot federated learning}, where a central server learns a global model over a network of federated devices in a single round of communication. Our approach---drawing on ensemble learning and knowledge aggregation---achieves an average relative gain of 51.5\% in AUC over local baselines and comes within 90.1\% of the (unattainable) global ideal. We discuss these methods and identify several promising directions of future work.
\end{abstract}
\section{Introduction}
Federated learning involves training models in a distributed fashion across a large network of IoT devices, such as mobile phones. Despite the known bottleneck of communication in this setting, current federated learning methods train models using iterative optimization techniques that require numerous rounds of communication between the devices and the central server~\cite[e.g.,][]{fedavg,mocha,konevcny2016federated}. 

In this work, we instead focus on techniques for \textit{one-shot federated learning}, in which we learn a global model from data in the network using only a single round of communication between the devices and the central server. We explore our proposed techniques in two common settings: (1) traditional supervised learning, in which each device generates its own set of training points with associated labels, and (2) semi-supervised learning, in which in addition to supervised device data, the central server has access to some relevant unlabeled data. 

Our key insight is that if each local device trains a local model to completion (as opposed to computing incremental updates as in traditional federated learning methods), we can effectively apply ensemble methods to capture global information across the device-specific models. To control client-to-server communication, we can restrict which local models are sent to the server using various protocols, e.g., via random sampling or by applying thresholds based on the amount of local data or local validation error. Additionally, if the server has access to a sample of unlabeled data, we can distill the resulting model ensembles into smaller, more concise models to control server-to-client communication.

We illustrate that these ideas can be used effectively and efficiently in large networks with thousands of devices, and demonstrate the efficacy of our approaches through a preliminary study on
real-world federated datasets. On average, our proposed approaches achieve a relative gain of 51.5\% in ROC-AUC over local baselines and come within 90.1\% of the (unattainable) ideal global model.

\section{Related Work} 

In federated learning, the aim is to train machine learning models directly across a network of IoT devices. Several challenges make this setting markedly different than learning in typical distributed settings: (1) communication is a critical bottleneck when learning across thousands to millions of IoT devices; (2) privacy is often a key concern, necessitating raw data to remain local; (3) variability across devices is profoundly different than in distributed data centers, as each device collects data in a non-IID fashion, and may have differing storage, computational, and communication capacities. While numerous recent methods have been proposed to address challenges (1)-(3), current approaches rely on \textit{iterative optimization techniques} to learn a global model---continually communicating updates to and from the central server and the local devices until convergence is reached~\cite{fedavg,mocha,konevcny2016federated}.

Instead, we propose that one-shot learning is an attractive approach for communication-efficient federated learning. While simple one-shot schemes, such as parameter averaging, have been explored in the distributed setting, it is well-known that the mean squared error for ERM problems solved in this fashion tends to decrease as $O(N^{-1/2} + \frac{m}{N})$, where $m$ is the number of machines and $N$ the total number of samples, thus requiring $m \leq \sqrt{N}$ to match performance of centralized ERM~\cite{zhang2012communication}. 

In the federated setting, where there may be a large number of devices $m$, many of which have very few local data points, we instead hypothesize that \textit{ensemble learning} techniques are better-suited for global modeling than naive averaging. Additionally, we note that ideas such as simple averaging can become complicated or infeasible for models such as deep learning or kernelized SVMs, which can require aggregating across differing architectures or between disparate sets of dual variables. While ensemble learning is commonly used to combine multiple learners for improved predictive performance, one challenge in the federated setting is that the final ensemble model could grow quite large, as there may be thousands to millions of devices in the network, each one generating a local model that we could potentially consider in our ensemble. We propose strategies for selecting a subset of local models in the resulting ensemble in Section~\ref{sec:methods}.

Finally, we explore a common setting, in which, in addition to the federated data, the central server may have access to some unlabeled proxy data. For example, it may be that a certain number of devices agree to share their local data, or that there exists some prior publicly available data for the problem at hand~\cite{jain2013differentially}. In this semi-supervised setting, we leverage \textit{distillation} \citep{hinton2015distilling} to further reduce the size of the resulting global model and potentially provide additional privacy guarantees, e.g., for models such as dual SVMs that would otherwise require sending raw data from each device.  

\vspace{-.5em}
\section{Methods}
\label{sec:methods}
We present ensemble methods for the on-device supervised setting, and extend these with distillation in the semi-supervised setting. Our presentation focuses on convex models (kernelized support vector machines) for binary classification tasks. However, the discussed methods are quite general-purpose, and may be easily extended to non-convex models (e.g., deep networks).

\textbf{Ensemble (Supervised)}:  For $m$  devices, each device $t \in [m]$ possesses local data $X_t \in \mathbb{R}^{d \times n_t}$ and solves: 
\begin{align}
\min_{w_t \in \mathbb{R}^d} \bigg\{ \mathcal{P}(w_t) := \dfrac{1}{n_t}\sum_{i=1}^{n_t} \ell_i(x_i^Tw_t) + \dfrac{\lambda}{2}\vert\vert w_t\vert\vert^2\bigg\}
\end{align}

where the vectors $\{x_i\}_{i=1}^{n_t} \in X_t$ and $\ell_i$ are real-valued convex loss functions (i.e. hinge loss). In the kernelized setting, we solve the dual formulation of this problem:

\begin{align}
\max_{\alpha_t \in \mathbb{R}^{n_t}} \bigg\{-\dfrac{1}{2\lambda n_t^2}\alpha_t^T\phi(X_t)^T \phi(X_t)\alpha_t+ \dfrac{1}{n_t}\sum_{i=1}^{n_t} -\ell_i^*(-{[\alpha_t]}_i)\bigg\}
\end{align}

where $\phi(X_t)^T \phi(X_t)$ can be replaced with $k(X_t, X_t)$ using the "kernel trick". All devices use an RBF kernel. We denote the local model learned on device $t$ as $f_t$, where $f_t(x) = w_t^Tx$. 
Upon completion, devices send local $f_t$ to the central server. Given $f_1, ..., f_m$ from devices in the network, the central server curates an ensemble of $k \leq m$ models. As the quality of local models may vary drastically (stemming from disparities in the distribution of data across devices), the optimal ensemble may consist of local models from only some devices. We discuss several strategies for ensemble selection:
\begin{enumerate}
\item  Cross-Validation (CV) Selection: Devices only share their local models if they achieve some baseline  performance (e.g., in terms of ROC AUC) on their local validation data, with the baselines determined in advance by the server. The server ensembles the $k$ best performing models from this subset of local models.   
\item Data Selection: Devices only share their local models if they have some baseline amount of local training data, with the baseline determined in advance by the server. The server ensembles models from these local models trained on the top $k$ largest data sets. 
\item Random Selection: The server randomly selects $k$ devices from the network and creates an ensemble from the corresponding local models. 
\end{enumerate}
The final ensemble $F_k$ of $k$ device models is constructed by averaging the predictions of each model. 

\textbf{Distillation (Semi-Supervised)}: When $k$ is large, communicating $F_k$ to each device (and performing inference) may be infeasible. When the central server has access to unlabelled public proxy data, $F_k$ can be compressed into a smaller model, $f'$, via distillation. In traditional distillation, knowledge from a "teacher" model is transferred to a "student" model by training the student on data labelled with the class probabilities output by the teacher network \cite{hinton2015distilling}. We present a modified approach adapted for binary classification with SVMs. For proxy data $x'_1, ... x'_l$, we generate corresponding "soft" labels $F_k(x'_1), ... ,F_k(x'_l)$. In particular, we perform distillation in the dual by minimizing the L2 difference in predictions between the student and teacher on the proxy data:

\begin{align}
\min_{\alpha' \in \mathbb{R}^l}\dfrac{1}{l} \sum_{i=1}^l \big(F(x'_i) - \sum_{j=1}^l \alpha'_j k(x'_j, x'_i)\big)^2
\end{align}

to produce a distilled model $f'(x)$=$\sum_{i=1}^l \alpha_i \phi(x'_i)$. When there are privacy concerns with sharing local models between devices (e.g., for dual SVMs, which require local support vectors to be shared), distillation not only helps to compress the model but also enables privacy-preserving learning.

\begin{figure}[t!]
    \centering
    \makebox[\textwidth]{
        \includegraphics[width=6cm]{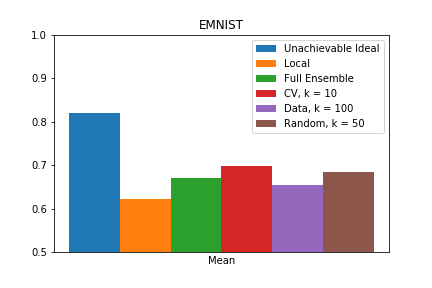}\hskip -3ex
        \includegraphics[width=6cm]{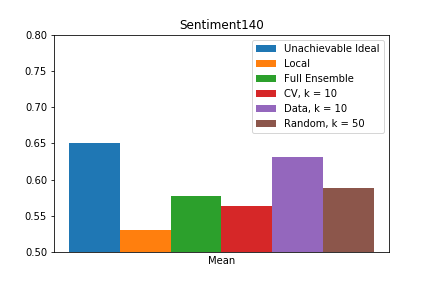}\hskip -3ex
        \includegraphics[width=6cm ]{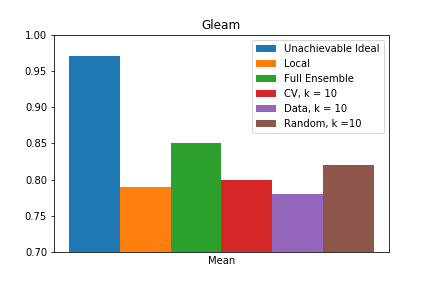}
    }
    \caption{Mean AUC across devices. Ensemble methods consistently outperform the local baseline.}
    \label{fig:means}
\end{figure}

\vspace{-.5em}
 \begin{figure}[h!]
\hskip -3.5ex
  \begin{minipage}[]{0.6\textwidth}
    \centering
    \begin{tabular}{lcc}
    \toprule
    \bf Dataset & \bf Total samples (devices) & \bf Device Min/Max  \\ \midrule
    EMNIST & 406,048 (3,462) & 10/460  \\ \midrule 
    Sent140 & 161,966 (4,000) & 21/345  \\ \midrule
    Gleam & 2,469 (38) & 33/99  \\ \bottomrule
    \end{tabular}
    \captionsetup{justification=centering}
    \captionof{table}{Summary of federated datasets for empirical study.}
    \label{table:datasets}
    \end{minipage}
    \hskip 5ex
    \begin{minipage}[]{0.39\textwidth}
    \centering
    \includegraphics[width=6cm]{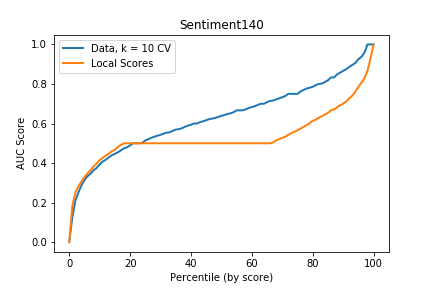}
    \captionof{figure}{Sent140 device score distribution. Ensemble methods significantly improve over local models.}
    \label{fig:sent140_perc}
  \end{minipage}
 \hskip -5ex
\end{figure}
\vspace{-2.5em}

\section{Results}

We test our methods on the following tasks/datasets, which have been explored in previous work on federated learning \cite{DBLP:journals/corr/abs-1812-01097, mocha}. Full details are provided in Table \ref{table:datasets}.
\begin{itemize}
    \item \textbf{EMNIST} \footnote{\url{https://www.nist.gov/itl/iad/image-group/emnist-dataset}}: Handwritten characters authored by different individuals (devices). We predict between lowercase and uppercase letters.
    \item \textbf{Sentiment140} \footnote{\url{http://help.sentiment140.com/for-students}}: Binary sentiment detection on tweets from different users (devices). We use a bag-of-words representation to featurize tweets.  
    \item \textbf{Gleam} \footnote{\url{http://www.skleinberg.org/data/GLEAM.tar.gz}}: Two hours of high resolution Google Glass sensor data corresponding to different activities. We predict between eating and other activities (walking, talking, etc). 
\end{itemize}

We split each device's local data into a 50/40/10 train-test-validation split. In constructing ensembles, we only consider classifiers from devices with a minimum number of local samples (30 for Gleam/Sent140, 60 for EMNIST). Devices with fewer data points are unlikely to learn informative local models. Enforcing this threshold simplifies ensemble construction for the central server, reduces communication required, and eases workload on data deficient devices.  We select ensembles (described in Section 3) for $k = 1, 10, 50, 100$, and evaluate against two baselines: 
\begin{itemize}
    \item \textbf{Unachievable\footnote{We are not aware of any  method (aside from solving an approximate primal problem that relies on a random feature-based approach) to solve a RBF-kernelized SVM in the federated setting while keeping all data local.} Ideal}: A "global" classifier, trained on data aggregated across all agents. This potentially violates both communication and privacy constraints. 
    \item \textbf{Fully Local Classifier}: A model learned only on a device's local data. For data deficient devices ($<30$ points for Gleam/Sent140, $<60$ points for EMNIST) we learn a constant classifier.    
\end{itemize}
Figure \ref{fig:means} compares the mean AUC (across devices) for the best $k$ on each selection strategy, the baseline approaches, and a full ensemble consisting of all device models. For the random ensemble, we report the average of 5 different trials. We find that ensemble approaches outperform the local baseline, and, with the exception of Gleam (which has relatively few devices), selected ensembles outperform full ensembles. Analyzing the distribution of device scores for Sent140 (Figure \ref{fig:sent140_perc}), we see that ensemble methods match high performing local models, while outperforming moderate-poor local models.  
 
\begin{figure}
    \centering
    \noindent
    \makebox[\textwidth]{
        \includegraphics[width=6cm]{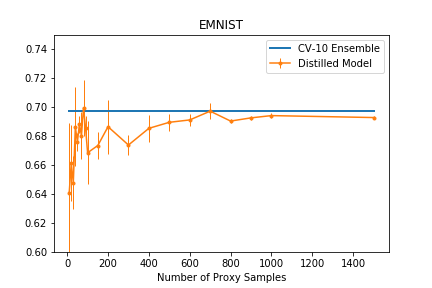}\hskip -3ex
        \includegraphics[width=6cm]{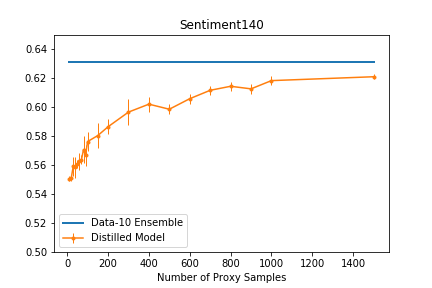}\hskip -3ex
        \includegraphics[width=6cm]{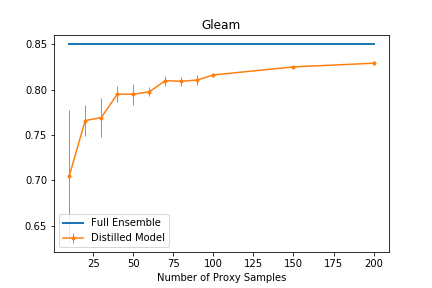}
    }
    \caption{Comparison of distilled model and ensemble for different proxy data sizes (averaged over 5 trials). The distilled model matches the original ensemble performance with relatively few proxy samples.}
    \label{fig:dist}
    \vspace{1em}
\end{figure}

 In the semi-supervised setting, we generate proxy data by sampling validation data across all devices. For each dataset, we distill the best performing ensemble and compare the distilled model to the ensemble as the size of proxy data increases (Figure \ref{fig:dist}). We find the distilled model can approximately match the original ensemble performance with a relatively small number of proxy samples. 

\section{Conclusion and Future Work}
Our work constitutes a preliminary investigation into one-shot federated learning. 
Our experiments indicate that ensemble methods and distillation---both unexplored in federated settings---yield promising results and suggest interesting future directions. These  include: (1) identifying "cohorts" of devices with similar local data distributions (e.g. devices from the same geographic region), which would allow us to learn ensembles that we could personalize for each device, (2) exploring the formal privacy guarantees of distillation in federated settings \cite{papernot2016semi}, (3) improving accuracy by moving from one-shot to few-shot federated learning, and (4) exploring our approaches in the context of non-convex models (e.g., deep neural networks). 

\subsection*{Acknowledgements}
This work was supported in part by DARPA FA875017C0141, the National Science Foundation
grants IIS1705121 and IIS1838017, an Okawa Grant, a Google Faculty Award, an Amazon Web
Services Award, a Carnegie Bosch Institute Research Award, and the CONIX Research Center. Any opinions, findings and conclusions or recommendations expressed in this material are those of the author(s) and do not necessarily reflect the views of DARPA, the National Science Foundation, or any other funding agency.

\bibliographystyle{plain}
\bibliography{sources}

\begin{thebibliography}{1}

\bibitem{DBLP:journals/corr/abs-1812-01097}
Sebastian Caldas, Peter Wu, Tian Li, Jakub Konecn{\'{y}}, H.~Brendan McMahan,
  Virginia Smith, and Ameet Talwalkar.
\newblock {LEAF:} {A} benchmark for federated settings.
\newblock {\em CoRR}, abs/1812.01097, 2018.

\bibitem{hinton2015distilling}
G.~Hinton, O.~Vinyals, and J.~Dean.
\newblock Distilling the knowledge in a neural network.
\newblock {\em arXiv preprint arXiv:1503.02531}, 2015.

\bibitem{jain2013differentially}
P.~Jain and A.~Thakurta.
\newblock Differentially private learning with kernels.
\newblock In {\em ICML}, 2013.

\bibitem{konevcny2016federated}
J.~Kone{\v{c}}n{\`y}, B.~McMahan, F.~Yu, P.~Richt{\'a}rik, A.~Suresh, and
  D.~Bacon.
\newblock Federated learning: Strategies for improving communication
  efficiency.
\newblock {\em arXiv preprint arXiv:1610.05492}, 2016.

\bibitem{fedavg}
B.~McMahan, E.~Moore, D.~Ramage, S.~Hampson, and B.~Aguera y~Arcas.
\newblock Communication-efficient learning of deep networks from decentralized
  data.
\newblock In {\em AISTATS}, 2017.

\bibitem{papernot2016semi}
N.~Papernot, M.~Abadi, U.~Erlingsson, I.~Goodfellow, and K.~Talwar.
\newblock Semi-supervised knowledge transfer for deep learning from private
  training data.
\newblock {\em arXiv preprint arXiv:1610.05755}, 2016.

\bibitem{mocha}
Virginia Smith, Chao-Kai Chiang, Maziar Sanjabi, and Ameet Talwalkar.
\newblock Federated multi-task learning.
\newblock In {\em Neural Information Processing Systems}, 2017.

\bibitem{zhang2012communication}
Y.~Zhang, M.~Wainwright, and J.~Duchi.
\newblock Communication-efficient algorithms for statistical optimization.
\newblock In {\em Neural Information Processing Systems}, 2012.

\end{thebibliography}
\end{document}